%% file: neurips_2026.tex
\documentclass{article}

 \usepackage[preprint]{neurips_2026}

\usepackage[utf8]{inputenc} 
\usepackage[T1]{fontenc}    
\usepackage{hyperref}       
\usepackage{url}            
\usepackage{booktabs}       
\usepackage{amsfonts}       
\usepackage{nicefrac}       
\usepackage{microtype}      
\usepackage{xcolor}         

\usepackage{graphicx}
\usepackage{amsmath}
\usepackage{wrapfig}

\usepackage{mathtools}
\usepackage{booktabs}

\usepackage{amssymb}
\usepackage{bm}
\usepackage{multirow}
\usepackage{subcaption}
\usepackage{pgfplots}
\usepackage{tikz}
\usepackage{algorithm}
\usepackage{wrapfig}
\usepackage{float} 
\usepackage[noend]{algpseudocode}  

\title{Quantifying Potential Observation Missingness in Inverse Reinforcement Learning}

\author{%
  Leo Benac \\
  School of Engineering and Applied Sciences\\
  Harvard University\\
  \texttt{lbenac@g.harvard.edu} \\
  \And
  Abhishek Sharma \\
  School of Engineering and Applied Sciences\\
  Harvard University\\
  \texttt{abhisheksharma@g.harvard.edu} 
  \AND
  Alihan Hüyük \\
  School of Engineering and Applied Sciences\\
  Harvard University\\
  \texttt{ahuyuk@fas.harvard.edu} 
  \And
  Finale Doshi-Velez \\
  School of Engineering and Applied Sciences\\
  Harvard University\\
  \texttt{finale@seas.harvard.edu} \\
}

\begin{document}

\maketitle

\begin{abstract}
    Inverse reinforcement learning (IRL), which infers reward functions from demonstrations, is a valuable tool for modeling and understanding decision-making behavior. Many variants of IRL have been developed to capture complexities of human decision-making, such as subjective beliefs, imperfect planning, and dynamic goals. However, an often-overlooked issue in real-world behavioral datasets is that the recorded data may be missing observations that were available to the original decision-maker. In use-inspired settings such as healthcare, this can make expert actions appear suboptimal, even when they were near-optimal given the information available at the time. As a result, the rewards learned by standard IRL may be misleading. In this paper, we identify the minimal perturbations to the recorded observations needed for the expert's actions to appear optimal. We develop a practical algorithm for this problem and demonstrate its utility for quantifying the possible extent of missing observations in behavioral datasets through extensive experiments on synthetic navigation tasks, a cancer treatment simulator, and ICU treatment data.
\end{abstract}

\vspace{-0.3cm}
\section{Introduction}
\vspace{-0.3cm}
A common use of inverse reinforcement learning (IRL) in healthcare is to analyze retrospective treatment data: given trajectories of patient vital signs (states) and clinician treatments (actions), can we infer the implicit objectives guiding those decisions? More generally, IRL aims to determine what reward function a decision-making agent might be optimizing given observations of their past actions \citep{ng2000algorithms}. Although IRL is typically performed as an intermediary step in \textit{imitation learning} (mimicking the decision-making policy of some demonstrator) \citep[e.g.][]{brown2020safe,ruan2023causal} or \textit{apprenticeship learning} (matching the demonstrator's performance in terms of some notion of ground-truth rewards) \citep[e.g.][]{abbeel2004apprenticeship}, it has also been an invaluable tool for modeling and understanding human decision-making behavior---providing a means to efficiently infer the likely goals of expert decision-makers and concisely describe them as reward functions.

In line with this \textit{descriptive purpose}, the literature on IRL has focused on accounting for various complexities of human decision-making that are not typically encountered in reinforcement learning---creating reward-based models of experts when they might have subjective beliefs about their environment that are different from the objective truth \citep{reddy2018you,huyuk2021explaining}, when their ability to plan future actions is not perfect \citep{jarrett2021inverse,poiani2024inverse}, or when their goals might be shifting over time \citep{likmeta2021dealing,huyuk2022inverse}. Additionally, recent descriptive models have leveraged this assumption of expert near-optimality to directly infer the environment's unknown transition dynamics rather than a reward function \citep{benac2024inverse}.

However, one complication of analyzing observed behavior that is often overlooked in IRL is the possibility of missing observations---that is some of the information observed by the original decision-maker might not have been recorded for us to observe.
\footnote{This is not to be confused with partial observations, where both the original decision-maker and us as researchers do not observe parts of the environment state.}

\begin{wrapfigure}{R}{0.42\textwidth} 
\centering
\includegraphics[width=\linewidth]{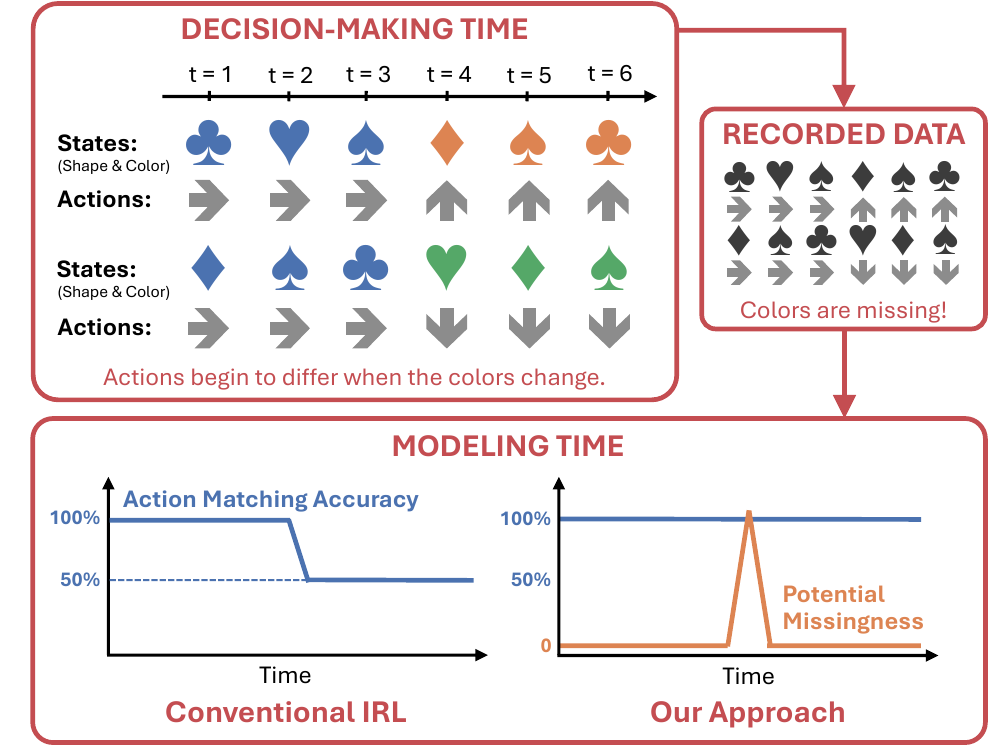}
\caption{
Suppose states consist of shapes and colors, and actions are determined solely by colors. The behavior in two episodes begins to differ after different color changes at $t=4$. If the recorded data omits colors, conventional IRL cannot accurately predict actions from that time onward. Our approach provides an alternative perspective: some unobserved change at $t=4$ is needed for the actions to be perfectly predictable.
}
\label{fig:intro}
\end{wrapfigure}

Missing observations can have significant implications, especially from a modeling perspective. For instance, consider a healthcare scenario where either Treatment~A or Treatment~B is most beneficial for a patient depending on some biomarker, say whether their blood pressure is low or high (suppose it is low 25\% of the time). Accordingly, doctors always assign Treatment~A to patients with low blood pressure and Treatment~B to patients with high blood pressure. Now suppose blood pressure is not recorded in a dataset of treatment decisions, despite being available to doctors at decision time. If we use conventional IRL with this dataset, we might conclude that Treatment~B has higher reward but that 25\% of the doctors act sub-optimally. However, an equally plausible explanation given our data is that the doctors act near-optimally, and that the factor making Treatment~A preferable is simply missing from the recorded observations. We might prefer one explanation over the other depending on our prior belief about the situation: the former if we trust the data collection process, or the latter if we trust the doctors' expertise, which is our assumption in this work. This type of retrospective analysis is representative of existing healthcare uses of IRL, including work on sepsis management, ICU hypotension management, and ICU ventilation and sedation decisions \citep{lee2019truly,srinivasan2020interpretable,yu2019inverse}. We provide a toy example in Figure~\ref{fig:intro} to convey the intuition of our setup.



Motivated by this gap in decision modeling with IRL, we ask the following question:
\textit{What is the smallest perturbation that needs to be made to observations so that the corresponding actions appear to be optimal?}
Of course, making this question meaningful requires a careful notion of what counts as a \emph{small} perturbation, which is one of the key technical issues we address. While answering this question does not allow us to recover the missing information itself, it can help us gauge how much information may be missing---more precisely, the minimal extent of missingness required if the decision-makers are assumed to be experts acting optimally. Determining whether this potential missingness is substantial can inform downstream choices about how the dataset should be used. For instance, if we intend to perform policy evaluation or regular reinforcement learning using a behavioral dataset, we might prefer algorithms designed to mitigate missing observations \citep[e.g.][]{kallus2020confounding,wang2021provably}.

\looseness=-1
\textbf{Contributions.} Our contributions are three-fold. \textit{Conceptually,} we introduce the idea of quantifying missingness in terms of minimal perturbations needed to align the actions of an expert with those of an optimal policy. We formulate this as a new optimization problem in Section~\ref{sec:contribution1}. \textit{Technically,} we develop an algorithm that can solve this optimization problem effectively (Section~\ref{sec:contribution2}). \textit{Empirically,} we demonstrate that we can recover perturbations with magnitude proportional to missingness, in a way that is robust to missingness in unimportant features, and that the learned perturbations capture meaningful structure in behavior by separating trajectories according to the underlying hidden decision context.

\section{Related Work}

\textbf{Missingness in Imitation Learning.}
As we briefly mentioned in the introduction, IRL is often performed as an intermediary step in imitation learning, which aims to copy the decision-making policy of a demonstrator. This is achieved by first inferring the demonstrator's reward function via IRL, and then optimizing that reward function via regular reinforcement learning to obtain their policy. Besides this strategy, other methods have also been developed for imitation learning that avoid inferring reward functions all together \citep[e.g.][]{ho2016generative}.

For reward-free imitation, \citet{zhang2020causal,kumor2021sequential} study missing observations from a causal perspective. They characterize when the demonstrator's policy can still be imitated despite missing variables in the underlying causal structure, i.e.\ despite hidden confounding. More recently, \citet{ruan2023causal} extended this line of work to reward-based imitation. Our goal is complementary. Rather than asking when imitation or reward recovery remains possible under hidden confounding, we ask how much unobserved information would be needed for the observed behavior in a fixed dataset to be explained as optimal under a reward model. This provides a quantitative measure of potential missingness, and can localize where missing context matters, without requiring a causal graph or assumptions on the underlying causal mechanisms.


\looseness=-1
\textbf{Modeling Variations in Behavior.}
As we discussed in the introduction, when observations are missing, behavior might seem like it is variable across different episodes, whereas actually, these variations can be explained perfectly through the variation of those missing observations. In our earlier healthcare example, the data we recorded made it seem like different doctors assign different treatments, however, these variations were simply due to differences in patient blood pressures. In that sense, our method can be viewed as a way of explaining variations that occur specifically across episodes, and as such, it is related to other methods that aim to explain variations in behavior using IRL.

Multi-modal IRL \citep{wang2017robust,hsiao2019learning,myers2022learning,qiao2024multi} and multi-task IRL \citep{babes2011apprenticeship,choi2012nonparametric,ramponi2020truly,huang2021driving} also explain variation in observed behavior, typically by inferring $K$ latent modes, decision-makers, or reward functions. Our setting is related but asks a different question: rather than introducing multiple experts or rewards, we ask how much heterogeneity can be explained by missing observations while retaining a shared reward function. Thus, these methods vary the number of latent modes, whereas we quantify the minimal unobserved context needed to make the observed actions appear optimal.

Finally, econometrics and statistics have long studied sensitivity to missing variables through omitted-variable-bias analyses for regression and treatment-effect estimation. Prior work uses selection on observables to reason about selection on unobservables \citep{altonji2005selection}, coefficient stability and $R^2$ bounds to assess unobserved confounding \citep{oster2019unobservable}, and robustness values or partial-$R^2$ summaries computable from standard regression outputs \citep{cinelli2020making}. Our work is in the same spirit, but for sequential decision-making: instead of asking whether a scalar coefficient is robust to an omitted variable, we ask how much unobserved trajectory-level information is needed for observed decisions to be consistent with optimal behavior under a reward model.

\section{Problem Formulation}
\vspace{-0.30cm}
\label{sec:contribution1}
\textbf{Setting.}
We consider a setting where agents act through $N$ episodes, indexed by $n\in[N]$, for $T$ time steps each, indexed by $t\in[T]$. At each time step, they encounter a state $s_{nt}\in \mathcal{S}$ and take an action $a_{nt}\in \mathcal{A}$ based on that state according to some behavior policy: $a_{nt}\sim\pi_{\text{behavior}}(s_{nt})$, assumed to be near-optimal. Now, suppose each state consists of two parts: $s_{nt}=(x_{nt},\allowbreak\tilde{x}_{nt})\in \mathcal{S}=\mathcal{X}\times\tilde{\mathcal{X}}$. While the agents determine their actions based on both parts, $s_{nt}$, suppose only the first part, $x_{nt}$, is recorded in a dataset while the latter part, $\tilde{x}_{nt}$, goes unrecorded, resulting in a behavioral dataset $\bm{D}=\{x_{nt},\allowbreak a_{nt}\}\in(\mathcal{X}\times \mathcal{A})^{N\times T}$ with missing observations $\{\tilde{x}_{nt}\}$.

\textbf{Objective.}
Our objective is to obtain a reward-based description of the behavior policy $\pi_{\text{behavior}}$ while accounting for the missing observations $\{\tilde{x}_{nt}\}$. In general, we cannot recover these missing observations, so we do \textit{not} aim for imputation; nor do we aim to replicate the behavior policy, so we do \textit{not} aim for imitation. Instead, we quantify how large the missing information might be under the prior belief that $\pi_{\text{behavior}}$ is near-optimal---for example, because it reflects decisions by domain experts such as clinicians. This provides a way to assess the dataset~$\bm{D}$ and inform downstream use: large potential missingness may suggest improving data collection or treating downstream analyses with caution, while small potential missingness supports greater confidence in tasks such as off-policy evaluation or offline reinforcement learning.

\textbf{Conventional IRL.}
As a contrast, consider applying conventional IRL to the recorded observations alone, ignoring the missing observations $\{\tilde{x}_{nt}\}$. Given a reward function $r_{\theta}(x,a)$ with corresponding optimal policy $\pi^*_{r_{\theta}}$, IRL seeks a reward under which the recorded actions appear as optimal:
\begin{align}
    \textstyle \textit{minimize}_{\theta} ~ \sum_{n,t}\|a_{nt}-\pi_{r_{\theta}}^*(x_{nt})\| .
\end{align}
where $\|\cdot\|$ is a distance measure between the actions of the experts and the inferred policy's decisions. For instance, in the case of maximum entropy IRL \citep{ziebart2008maximum}, this would be the negative log-likelihood of actions under the inferred policy: $\|a_{nt}-\pi_{r_{\theta}}^*(x_{nt})\|\doteq -\log\pi_{r_{\theta}}^*(x_{nt})[a_{nt}]$.

It is important to note that a good match between actions and the inferred policy may not always be possible. In particular, if the same recorded observation $x_{nt}=x_{n't'}$ has led to different actions $a_{nt}\neq a_{n't'}$, then any policy of the form $\pi^*_{r_{\theta}}(x)$ must assign the same decision to both inputs and therefore cannot match both actions simultaneously. This is exactly the type of inconsistency that can arise with missing observations: the difference in actions may be due to differences in the unrecorded part of the state, $\tilde{x}_{nt}\neq\tilde{x}_{n't'}$, even though the recorded part is the same.

\textbf{Modeling Missingness.}
Recall that our goal is to reason about the information content of the missing observations $\{\tilde{x}_{nt}\}$, rather than to reconstruct each missing value itself. To avoid explicitly modeling the dynamics of the missing observations, we introduce for each trajectory $n\in[N]$ a static latent factor $z_n\in\mathcal{Z}$ that stands in for all missing information relevant to that trajectory. Here, $z_n$ may in general be a vector, and we do not assume that its dimension or structure is known a priori. It is \emph{static} only in the sense that it does not vary across time within a trajectory, hence it carries no time index. From this point on, we model rewards and policies as functions of the recorded observations together with this latent factor: $r_{\theta}(z_n,x_{nt},a_{nt})$ and $
a_{nt}\sim \pi(z_n,x_{nt})$.

This reformulation does not reduce expressiveness as long as $z_n$ can encode information equivalent to the full missing trajectory $(\tilde{x}_{n1},\tilde{x}_{n2},\ldots,\tilde{x}_{nT})$. We emphasize that $z_n$ is introduced as a compact representation of potentially missing context, not as an assumption that the true missing information is itself low-dimensional or static.

Then our objective can be stated as finding the smallest possible $\{z_n\}$ that allows a desired level of optimality, measured as the distance between recorded actions and an optimal policy:
\begin{alignat}{3}
    \textit{minimize}_{\{z_n\},\theta} ~ &\textstyle \sum_n && \|z_n\| \label{eqn:objective-min} \\
    \textit{s.t.} ~ &\textstyle \sum_{n,t} && \|a_{nt}-\pi^*_{r_{\theta}}(z_n,x_{nt})\| \leq \zeta. \label{eqn:objective-st}
\end{alignat}

\looseness=-1
Here $\zeta$ controls the required action-matching level. When $z_n=0$ for all $n$, the formulation reduces to conventional IRL and can only achieve thresholds above
\[
\zeta_0 = \min_\theta \sum_{n,t}\|a_{nt}-\pi^*_{r_\theta}(x_{nt})\|.
\]
As $\zeta$ decreases below $\zeta_0$, nonzero $z_n$ must supply additional information. At the other extreme, $\zeta=0$ can be achieved by encoding the full action sequence in $z_n$, but our objective asks how much smaller a perturbation suffices.


The optimization in Equations~\eqref{eqn:objective-min}--\eqref{eqn:objective-st} is still abstract at this stage. To make the size of $z_n$ meaningful, we must make concrete modeling choices about both the space $\mathcal{Z}$ and the way $z_n$ enters the reward function. We now turn to these choices.

\looseness=-1
\textbf{Magnitude of Latent Factors.}
A key issue is that, unless we design the space of $z_n$ and the class of reward functions $r_{\theta}(z_n,x_{nt},a_{nt})$ carefully, the magnitude of $\|z_n\|$ may not be meaningful. As a simple example, suppose $\mathcal{Z}=\mathbb{R}$, $\mathcal{X}=\mathbb{R}$, $\mathcal{A}=\{0,1\}$, and the reward functions are linear:
\[
r_{\theta}(z_n,x_{nt},a_{nt}) = \phi_{a_{nt}} z_n + \psi_{a_{nt}} x_{nt},
\]
with parameters $\theta=\{\phi_0,\phi_1,\psi_0,\psi_1\}$. Then we can make $\{z_n\}$ arbitrarily small by making $\{\phi_0,\phi_1\}$ sufficiently large, without changing the reward function or the optimal policy. In such a parameterization, the size of $z_n$ reflects an arbitrary scaling choice rather than the amount of missing information.

\looseness=-1
\textbf{$\mathcal{Z}$ Representation.} To make the magnitude of $z_n$ meaningful, we restrict how missing information enters the reward. In this work, we consider continuous state spaces and parameterize episode-specific perturbations using a finite set of $K$ kernels placed in the observed state space. Specifically, we introduce kernel centers $\mu_1,\ldots,\mu_K\in\mathcal{X}$ and a bandwidth parameter $\sigma>0$, both shared across episodes. Each episode $n\in[N]$ then has its own action-specific kernel coefficients $ z_n\in\mathcal{Z}=\mathbb{R}^{K\times |\mathcal{A}|}.$
We define the reward function as:
\begin{align}
    r_{\theta}(z_n,x_{nt},a_{nt})
    =
    \bar{r}_{\theta}(x_{nt},a_{nt})
    +
    \sum_{k=1}^{K}
    \kappa_{\sigma}(x_{nt},\mu_k)\,(z_n)_{k,a_{nt}},
    \label{eqn:reward-decomposition}
\end{align}
where $\bar{r}_{\theta}$ is a shared base reward function (represented as a neural network with parameter $\theta$) and $\kappa_{\sigma}(x,\mu_k)$ measures the proximity of $x$ to the $k$-th kernel center. In our implementation, we use Gaussian radial basis function kernels:
$
 \kappa_{\sigma}(x,\mu_k)
    =
    \exp\!\left(
        -\frac{\|x-\mu_k\|_2^2}{2\sigma^2}
    \right). $

\looseness=-1
\textbf{Size of $z_n$.}
Under this parameterization, $z_n$ affects the reward only \emph{linearly}: each coefficient $(z_n)_{k,a}$ directly scales one localized perturbation pattern for action $a$. Thus, the size of $z_n$ is directly tied to the size of the reward perturbation it induces. We measure trajectory-level missingness by the $\ell_1$ norm $\|z_n\|_1$, and use $\|z_n\|=\|z_n\|_1$ in Equation~\eqref{eqn:objective-min}. Smaller $\|z_n\|_1$ corresponds to weaker episode-specific deviations from the shared reward function.

Importantly, $z_n$ is penalized once per trajectory, not once per time step. Thus, a single latent component can influence multiple decisions within the same trajectory without being counted multiple times, matching our interpretation of $z_n$ as trajectory-level missing context. More generally, this kernel parameterization avoids assigning an independent perturbation to every possible continuous observation; instead, it represents missing information through a small number of spatially localized perturbation patterns, smoothly interpolated across the observed state space.



\section{Minimum-Perturbation IRL}
\label{sec:contribution2}
We now describe the optimization used to learn the base-IRL reward and episode-specific perturbations. Our method builds on the maximum causal entropy IRL framework \citep{ziebart2010modeling}, in which the expert is modeled as following a soft-optimal policy induced by a soft $Q$-function:
\begin{equation}
    \pi(a | x) = \frac{\exp(Q(x,a))}{\sum_{a' \in \mathcal{A}} \exp(Q(x,a'))}.
\end{equation}
The corresponding soft Bellman equations are:
\begin{align}
    Q(x,a) = \mathbb{E}_{x' \sim T(\cdot|x,a)}\!\left[r(x,a)+\gamma V(x')\right],
    \qquad
V(x)=\log\sum_{a\in\mathcal{A}}\exp(Q(x,a)).
\end{align}
In our offline setting, we do not have access to the transition model $T$. We therefore enforce these equations only approximately on transitions in the dataset using a sample-based Bellman target and a penalized objective. This forces us to regularize the Q-function more heavily.

\textbf{Minimum-Perturbation IRL (MP-IRL).}
We optimize the model in two phases. Phase~1 performs standard IRL to extract the best \emph{base}-reward explanation based solely on what the recorded data tells us. Phase~2 then freezes this base reward model and infers the episode-specific perturbations $z_n$ needed to recover the remaining behavior. This structure is central to our approach: it ensures that the latent variables only explain what the observed data cannot, allowing us to accurately quantify the perturbations corresponding to the missing observations.

We write
\[
    \pi_{\phi}(a \mid x,z,\mu)
    \propto
    \exp\!\left(\frac{Q_{\phi}(x,z,\mu,a)}{\tau}\right)
\]
for the soft policy induced by the critic $Q_{\phi}$, where $\tau>0$ is a temperature parameter and $\mu=\{\mu_k\}_{k=1}^K$ are the shared kernel centers. We use $\tau=0.3$ throughout. Smaller values of $\tau$ make the policy closer to deterministic, but this does not necessarily by itself improve action matching or likelihood; it only sharpens the action probabilities implied by the critic. Since $\tau$ changes the scale of the perturbations needed to explain the same actions, we keep it fixed whenever comparing the learned size of $z$ across experiments. We also define the soft Bellman target:
\begin{align}
    y_{\phi,\theta}(x_{nt},z_n,\mu,a_{nt},x_{n,t+1})
    =
    r_{\theta}(z_n, x_{nt},a_{nt};\mu)
    +
    \gamma \log \sum_{a' \in \mathcal{A}}
    \exp\bigl(Q_{\bar{\phi}}(z_n, x_{n, t+1},a{'};\mu)\bigr)
\end{align}
where $Q_{\bar{\phi}}$ denotes the target Q-network for stability and with the understanding that terminal next states have zero continuation value. Throughout, $\ell_{\mathrm{SL1}}$ denotes the smooth-$\ell_1$ (Huber) loss.

\textbf{Decision regions and kernel initialization.}
Although we do not give a formal definition, it is useful to think in terms of \emph{decision regions}: parts of the observed state space where nearly identical recorded observations can lead to different expert actions because relevant context is missing. These are precisely the regions where we expect reward perturbations to be needed. We normalize all features before computing kernel distances so that variables with larger numerical scale do not dominate the RBF kernels. If prior knowledge about decision regions is available, the centers can be initialized there directly. Otherwise, we initialize them with $K$-means on the recorded observations so that they start on the support of the data. We deliberately choose $K$ somewhat large; if some centers are not useful, the sparsity penalty in phase~2 drives their corresponding coefficients in $z_n$ close to zero.

\textbf{Phase 1 (base IRL).}
We initialize $z_n \gets \mathbf{0}$ for all episodes. Since the perturbation then vanishes, the reward reduces to the shared base reward $\bar r_\theta$. We solve
\begin{align}
\min_{\phi,\theta} \quad
\mathcal{L}_{\mathrm{base}}(\phi,\theta)
&=
-\sum_{n=1}^{N}\sum_{t=1}^{T_n}
\log \pi_{\phi}(a_{nt}\mid x_{nt},\mathbf{0};\mu)
\nonumber\\
&\quad+
\lambda \sum_{n=1}^{N}\sum_{t=1}^{T_n}
\ell_{\mathrm{SL1}}
\!\left(
Q_{\phi}(\mathbf{0}, x_{nt},a_{nt}; \mu),
y_{\phi,\theta}(x_{nt},\mathbf{0},\mu,a_{nt},x_{n,t+1})
\right).
\label{eq:mpirl_phase1_compact}
\end{align}
Let $(\phi^{(1)},\theta^{(1)})$ denote the solution.

\textbf{Phase 2 (MP-IRL).}
We freeze the reward parameters at $\theta^{(1)}$, initialize the critic from phase~1 with $\phi^{(1)}$, and optimize the critic together with the episode-specific kernel coefficients $\bm z=\{z_n\}_{n=1}^N$ and the kernel centers $\mu=\{\mu_k\}_{k=1}^K$. Thus, phase~2 solves

\begin{align}
\min_{\phi,\bm z;\mu} \quad
\mathcal{L}_{\mathrm{MP}}(\phi,\bm z;\mu)
&=
-\sum_{n=1}^{N}\sum_{t=1}^{T_n}
\log \pi_{\phi}(a_{nt}\mid x_{nt},z_n;\mu)
\nonumber\\
&\quad+
\lambda \sum_{n=1}^{N}\sum_{t=1}^{T_n}
\ell_{\mathrm{SL1}}
\!\left(
Q_{\phi}(z_n, x_{nt},a_{nt}; \mu),
y_{\phi,\theta^{(1)}}(x_{nt},z_n,\mu,a_{nt},x_{n,t+1})
\right)
\nonumber\\
&\quad+
\alpha \sum_{n=1}^{N}
\left(
\sum_{k=1}^{K}\|z_{n,k,:}\|_2 + \|z_n\|_1
\right)
\nonumber\\
&\quad+
\beta \sum_{n=1}^{N}\sum_{t=1}^{T_n} \sum_{a' \in \mathcal{A}}
\ell_{\mathrm{SL1}}
\!\left(
Q_{\phi}(\mathbf{0}, x_{nt}, a'; \mu),
Q_{\phi^{(1)}}(\mathbf{0}, x_{nt}, a';\mu)
\right).
\label{eq:mpirl_phase2_compact}
\end{align}

The first term in both phases encourages action matching, while the second enforces soft Bellman consistency so that the learned policy remains tied to reward optimality rather than pure behavior cloning. In Phase 2, the sparsity and $\ell_1$ penalties implement the minimum-perturbation objective by encouraging localized, low-magnitude reward changes through $z_n$. The final term anchors the phase-2 critic to the phase-1 critic when $z=0$, so improvements over Phase 1 must be explained by nonzero perturbations rather than by changing the zero-perturbation policy. For $z_n$ to be meaningful as a missingness measure, the Bellman and anchor errors should remain small; otherwise the critic could absorb behavior-cloning-like changes. Algorithm~\ref{alg:mpirl} in Appendix~\ref{sec:algo_appendix} gives the full training procedure.

\section{Experimental Setup}
\label{sec:experimental_setup}

\textbf{Environments and data.}
We evaluate our method on five domains: three synthetic continuous navigation tasks, a cancer simulator based on \citet{yauney2018reinforcement}, and an ICU hypotension-management task using MIMIC-IV \citep{johnson2020mimic}. The three navigation tasks share the same continuous 2D geometry but differ in how hidden information affects branching: a \emph{single-decision} task, a \emph{two-decision dependent} task, and a \emph{two-decision independent} task. For the synthetic domains and the cancer simulator, we generate $500$ expert trajectories per environment. More details are deferred to Appendices \ref{app:continuous_navigation}, \ref{app:cancer_simulator}, and \ref{ICUenvironment}.

\textbf{Evaluation metrics.}
We quantify potential missingness by the average \textbf{size} of $z$:
$
\frac{1}{N}\sum_{n=1}^{N}\|z_n\|_1,
$
where $\|z_n\|_1$ is the elementwise $\ell_1$ norm of the trajectory-level perturbation. Since $z_n$ enters the reward linearly, this quantity measures the average magnitude of the reward perturbation needed beyond the base-IRL reward. We also report action-matching \textbf{accuracy} and the negative log-likelihood (\textbf{NLL}) of expert actions under the learned policy. For navigation tasks, decision regions are known by construction, so we additionally report accuracy inside and outside these regions; for other domains, we report overall accuracy and NLL.

\begin{figure}[t]
    \centering
\includegraphics[width=.70\textwidth, clip]{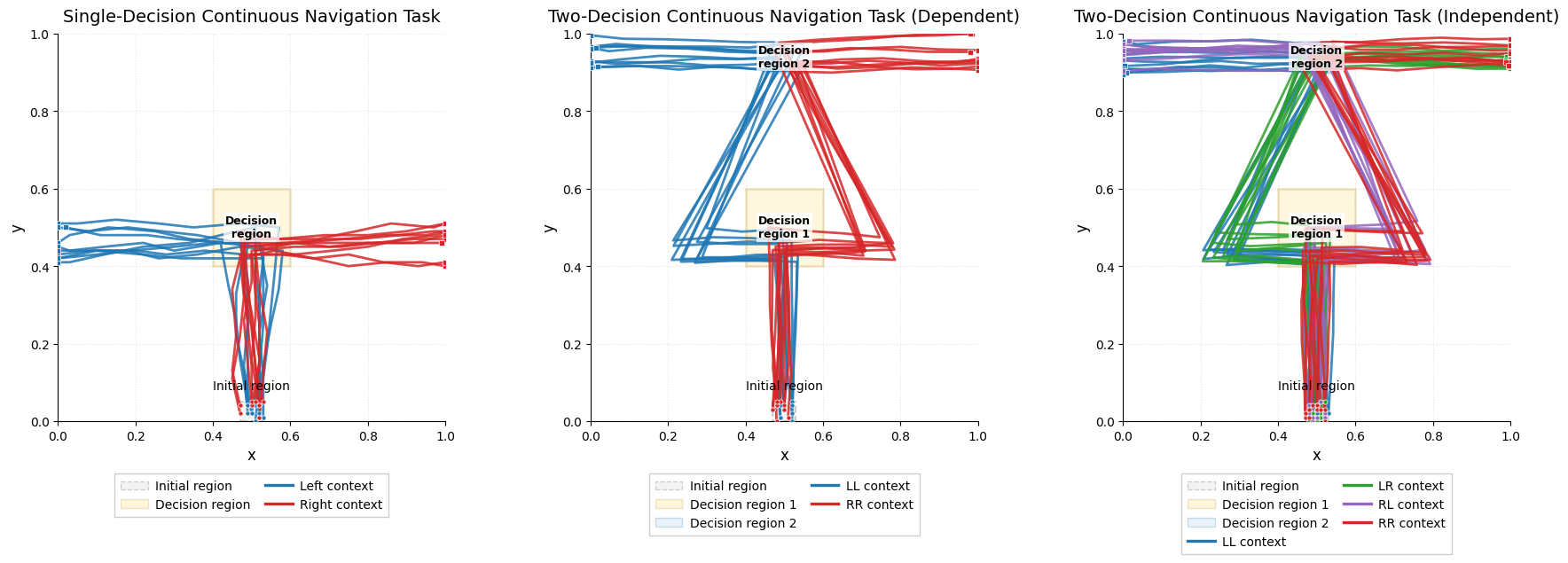}
\includegraphics[width=.65\textwidth]{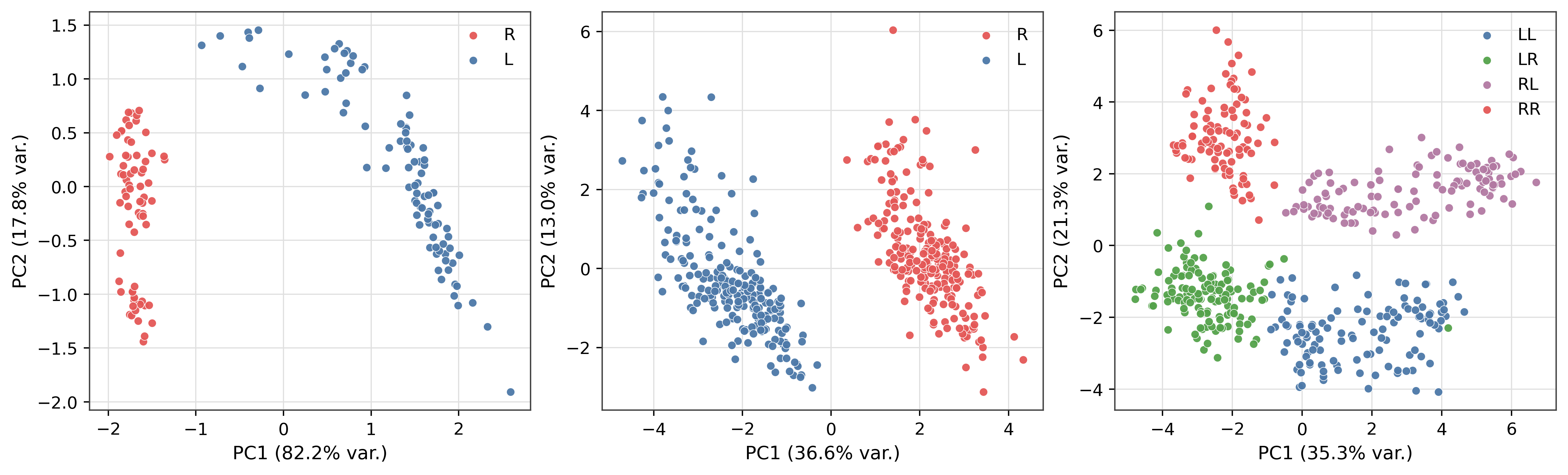}
    \caption{
\textbf{Continuous navigation tasks.}
Top: demonstrations; shaded boxes mark decision regions. Bottom: PCA projections of learned perturbations $z_n$, which separate trajectories by hidden context.
}
    \label{fig:synthetic_demos}
\end{figure}

\textbf{Continuous navigation tasks.}
The 3 navigation tasks are controlled settings where missing episode-level information affects behavior only in specific \emph{decision regions}. In all tasks, the learner observes only position $(x,y)$, while the expert also observes hidden context. In the single-decision task, a binary variable $z\in\{0,1\}$ determines whether the agent should go left or right after reaching a central region, so similar observed positions can require different actions. The two-decision tasks add a second branching point. In the dependent version, one hidden variable controls both decisions, yielding two trajectory families: left-left and right-right. In the independent version, two hidden variables $(z^{(1)},z^{(2)})$ control the decisions separately, yielding four families: left-left, left-right, right-left, and right-right. Thus, the independent task contains more missing information despite the same two decision regions. Example of expert demonstrations are shown in Figure~\ref{fig:synthetic_demos} for all 3 tasks.

\textbf{Cancer simulator.}
The cancer simulator is a low-grade glioma treatment-planning task with $T=30$ monthly decisions and a binary chemotherapy action ($a=1$ for treatment, $a=0$ otherwise). The full state contains drug concentration, proliferative tissue, quiescent tissue, damaged quiescent tissue, and time step. Expert trajectories, shown in Figure~\ref{fig:cancer_demos} of Appendix~\ref{app:cancer_simulator}, mainly reflect one patient type up to transition noise; time step already explains much of the treatment pattern. To study missingness, we evaluate three masks of the same expert dataset. From most to least missing information, the learner observes: (i) time step only; (ii) time step, quiescent tissue, and damaged quiescent tissue; and (iii) all features except quiescent tissue, the weakest predictor.

\textbf{ICU hypotension-management task.}
We evaluate on a real-life hypotension-management task (MIMIC-IV), where each trajectory is one ICU stay with hourly measurements and clinician treatments. States contain 14 clinical variables plus time, and actions are no treatment, vasopressor, IV fluid bolus, or both; preprocessing details are in Appendix~\ref{ICUenvironment}. We use three masks: time only, time plus low-predictive features, and all recorded features. The low-predictive mask tests whether $z$ size remains stable when added variables contain little decision-relevant information.


\section{Results}
\textbf{Continuous navigation tasks.}
The continuous navigation experiments test two questions: whether MP-IRL localizes where missing information is needed, and whether the learned perturbation size reflects the amount of missing information. Table~\ref{tab:full_results} shows that conventional IRL already matches the expert almost perfectly outside decision regions, with non-decision accuracy near $100\%$ in all three tasks. Its errors are concentrated inside decision regions, where similar observed position require different actions. In the single-decision task, conventional IRL achieves only $67.9\%$ decision-region accuracy and NLL $183.39$, while MP-IRL increases accuracy to $99.6\%$ and reduces NLL to $31.89$.

\begin{wrapfigure}{r}{0.47\textwidth}
    \centering
    \vspace{-0.8em}
\includegraphics[width=0.48\textwidth]{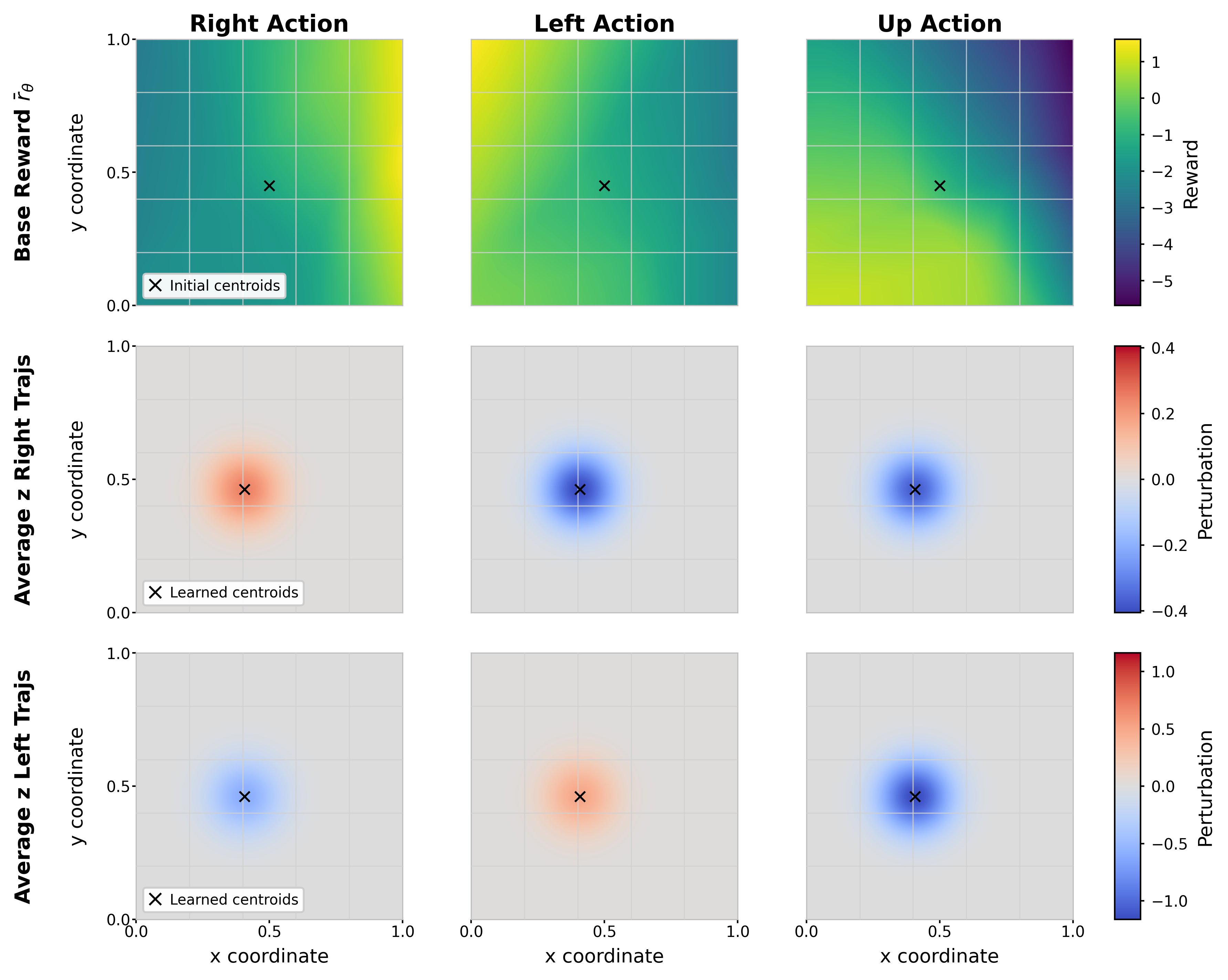}
    \vspace{-0.8em}
    \caption{Learned reward structure for the single-decision navigation task. The top row shows the base-IRL reward $\bar r_\theta$. The lower rows show the average learned perturbation $z$ for right- and left-context trajectories across right, left, and up actions; red indicates positive perturbations and blue negative perturbations. MP-IRL learns localized perturbations near the decision region that explain the expert behavior.}
    \label{fig:grid_1_reward}
    \vspace{-0.8em}
\end{wrapfigure}

Figure~\ref{fig:grid_1_reward} illustrates the mechanism in the single-decision task. The base reward explains behavior away from the center but cannot resolve the decision region, where hidden context determines whether the expert goes left or right. The learned perturbations add reward to the context-appropriate action near the decision region and penalize competing actions, so $z_n$ locally modifies the reward in the direction needed to explain the expert action. This visualization uses one kernel initialized in the decision region; other experiments use $K$-means initialization.

The two-decision tasks results show the same pattern (Table \ref{tab:full_results}). Conventional IRL performs well outside decision regions but poorly inside them, with decision-region accuracies of $59.6\%$ and $59.4\%$ for the dependent and independent tasks. MP-IRL raises these to $98.4\%$ and $97.1\%$, respectively, while substantially reducing NLL. The learned perturbations activate near decision regions and favor the expert actions; unused kernels receive near-zero coefficients due to the sparsity penalty.

The perturbation sizes also match the hidden structure. The single-decision and dependent two-decision tasks have similar $z$ sizes, $1.2$ and $1.1$, because both require one hidden bit. The independent task requires two hidden decisions, yields four trajectory families, and learns a larger perturbation size, $2.7$. Figures~\ref{fig:grid_R_ind} and~\ref{fig:grid_R_dep} show that perturbations concentrate near decision regions and favor the actions required by each hidden context. The PCA projections in Figure~\ref{fig:synthetic_demos} further show that $z_n$ separates into two groups for the single-decision and dependent tasks, and four groups for the independent task, matching the true hidden contexts.


\textbf{Cancer simulator.}
For the cancer simulator, we create missingness by removing features observable to the IRL agent. We evaluate three observation masks, ordered from least to most missing information: all features except quiescent tissue, damaged quiescent tissue plus quiescent tissue plus time, and time only. Example of demonstrations are shown in Figure~\ref{fig:cancer_demos}, where we notice time alone is already a strong feature for explaining most of the expert's actions. The more features are hidden, the more conventional IRL degrades: accuracy decreases from $93.73\%$ to $91.72\%$ and then $87.87\%$, while NLL increases from $1291.39$ to $1687.03$ and then $1988.29$ (see Table~\ref{tab:full_results}). MP-IRL recovers near-perfect accuracy in all settings, but the required perturbation size increases with missingness, from $0.54$ to $0.91$ and then $1.65$, consistent with interpreting $\|z\|_1$ as a measure of potential missingness.

\begin{figure}[h]
    \centering
\includegraphics[width=1.0\textwidth]{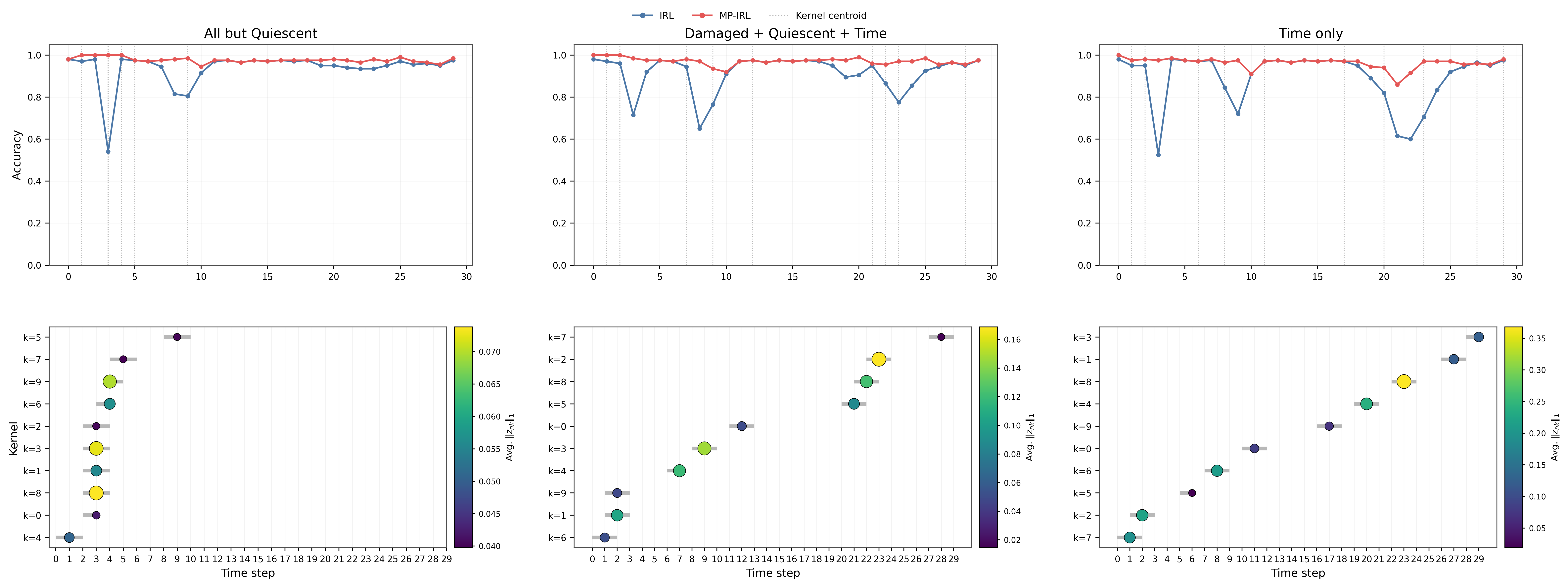}
    \caption{
\textbf{Cancer simulator results.}
Top: Accuracy over time. Bottom: Learned kernel centers, with marker proportional to average perturbation magnitude $\|z_{\cdot,k,:}\|_1$; gray bars show bandwidth.
}
\label{fig:cancer_main}
\end{figure}

We use $K=10$ kernels initialized with $K$-means in all cancer experiments. Figure~\ref{fig:cancer_main} shows the averaged accuracy over time (top row) and the learned kernel centers projected onto the time feature, along with the average perturbation magnitude $\|z_{\cdot,k,:}\|_1$ for each kernel (bottom row); gray horizontal bars indicate the kernel bandwidth. MP-IRL assigns larger perturbations to time points where conventional IRL has low accuracy. For example, in the time-only setting, conventional IRL fails around time steps $3$, $9$, and $22$, and MP-IRL places high-magnitude kernels near those regions. In the all-but-quiescent setting, the remaining ambiguity is mostly concentrated near time step $3$, where the learned kernels also concentrate. Thus, MP-IRL not only improves action matching, but also localizes where recorded observations are insufficient to explain expert treatment decisions, thereby learning meaningful minimal perturbations.

\begin{table}[t]
\centering
\scriptsize
\setlength{\tabcolsep}{3.2pt}
\renewcommand{\arraystretch}{0.95}
\begin{tabular}{llccc}
\toprule
Domain & Setting / mask & Acc. IRL$\to$MP-IRL & NLL IRL$\to$MP-IRL & Size $z$ \\
\midrule
\multirow{3}{*}{Nav.}
& Single-decision & $67.9{\to}99.6$ & $183.4{\to}31.9$ & $1.2$ \\
& Two-decision dep. & $59.6{\to}98.4$ & $1078.5{\to}304.9$ & $1.1$ \\
& Two-decision ind. & $59.4{\to}97.1$ & $1176.7{\to}269.4$ & $2.7$ \\
\midrule
\multirow{3}{*}{Cancer}
& All but quiescent & $93.73{\to}97.68$ & $1291.4{\to}477.4$ & $0.54$ \\
& Damaged + quiescent + time & $91.72{\to}97.17$ & $1687.0{\to}834.5$ & $0.91$ \\
& Time only & $87.87{\to}96.26$ & $1988.3{\to}1225.8$ & $1.65$ \\
\midrule
\multirow{3}{*}{ICU}
& All recorded features & $63.22{\to}89.41$ & $12407{\to}5020$ & $5.63$ \\
& Time + low-pred. features & $55.24{\to}90.66$ & $15299{\to}4795$ & $13.56$ \\
& Time only & $54.79{\to}90.21$ & $15372{\to}5033$ & $13.66$ \\
\bottomrule
\end{tabular}
\vspace{0.4em}
\caption{
\textbf{Compact main results.}
Accuracy is decision-region accuracy for navigation and overall accuracy otherwise.
Full results, including non-decision-region accuracy, are in Appendix~\ref{app:full_results}, Table~\ref{tab:full_results}.
}
\label{tab:main_results}
\vspace{-1em}
\end{table}

\textbf{ICU hypotension-management task.}
We run three experiments on the MIMIC-IV hypotension task using different observation masks: time step only, time step plus low-predictive features, and all recorded features. Because this is a real clinical dataset, even the full recorded state may omit information available to clinicians at decision time, such as bedside assessments or patient-specific context. We use $K=30$ kernels in all settings.

Table~\ref{tab:full_results} shows that the time-only and time-plus-low-predictive-feature settings have similar accuracy, NLL, and learned $z$ size. This shows that MP-IRL is robust to unimportant features: the perturbation size reflects missing decision-relevant information, not simply the number of observed features. When all recorded features are included, conventional IRL achieves higher accuracy and lower NLL, and MP-IRL requires a smaller perturbation, again supporting the interpretation of $z$ size as a measure of potential missingness.

Across all three masks, MP-IRL substantially improves over conventional IRL, although action-matching accuracy remains around $90\%$. This is expected in a noisy and heterogeneous clinical setting where decisions may depend on information absent from MIMIC-IV. Still, this accuracy is higher than prior model-based RL results on the same task, which reported action-matching accuracies around $50$--$60\%$ for the strongest methods \citep{benac2024inverse}. Overall, MP-IRL remains useful in this real-world setting: it improves action matching, is robust to low-predictive features, and learns smaller perturbations when more decision-relevant information is observed.

\section{Conclusion}
We introduced MP-IRL, a method for quantifying potential observation missingness in inverse reinforcement learning. The method asks how much trajectory-level information must be added to the recorded observations for expert actions to appear optimal under a reward model. Across continuous navigation tasks, MP-IRL localizes perturbations to decision regions and learns perturbation sizes that match the true amount of hidden context. In the cancer simulator and MIMIC-IV ICU hypotension task, the learned perturbation size decreases as more decision-relevant information is observed and remains stable when low-predictive features are added. These results support the use-inspired motivation of our work: in healthcare datasets, apparent expert suboptimality may reflect missing clinical context rather than poor decisions.

\textbf{Limitations.}
A limitation of MP-IRL is that the kernel-based perturbation representation may become difficult to learn effectively in high-dimensional state spaces, where placing and optimizing localized kernels requires substantially more data and careful regularization. In such settings, future work may need more scalable perturbation parameterizations while preserving the interpretability of the learned missingness measure.

\bibliographystyle{plainnat}
\bibliography{references}

\appendix

\section{Training Procedure}
\label{sec:algo_appendix}

\subsection{Algorithm}

\textbf{Minimum-Perturbation IRL algorithm.}
Algorithm~\ref{alg:mpirl} summarizes our training procedure. The method has a simple structure: it first runs conventional IRL on the recorded observations alone, and then asks how much additional episode-specific perturbation is needed to explain the remaining mismatch with expert behavior.

In \textbf{Phase~1 (Base IRL)}, we set all perturbations to zero and optimize the shared reward model and critic by minimizing Equation~\eqref{eq:mpirl_phase1_compact}. Since $z_n=0$ for all episodes, the reward reduces to the shared component $\bar r_\theta$. The resulting parameters $(\phi^{(1)},\theta^{(1)})$ therefore capture the part of the behavior that can already be explained from the recorded observations alone.

In \textbf{Phase~2 (Minimum-Perturbation IRL)}, we freeze the shared reward at $\theta^{(1)}$ and optimize the critic together with the episode-specific coefficients $\{z_n\}_{n=1}^N$ and, when desired, the kernel centers $\mu$. This is the key step of the method. Because the shared reward is fixed, the model can no longer improve action matching by changing the common reward structure across all trajectories. Instead, any additional improvement must come from nonzero episode-specific perturbations. In this sense, Phase~2 directly operationalizes the question posed in Section~\ref{sec:contribution1}: what is the smallest perturbation needed for the observed actions to appear optimal?

The minimum-perturbation aspect is enforced by the penalty term in Equation~\eqref{eq:mpirl_phase2_compact}, which encourages only a small number of kernels to become active and keeps their coefficients small. As discussed above, this is meaningful because $z_n$ enters the reward only as a linear perturbation, and the penalty is applied once per trajectory rather than once per time step. The critic anchoring term is also important: it keeps the phase-2 critic close to the phase-1 critic when $z=0$, preventing the critic from explaining the demonstrations better on its own without using the perturbation variables. As a result, the gains achieved in Phase~2 are attributable to the learned perturbations rather than to additional flexibility in the critic.

Finally, the kernel centers determine \emph{where} in the observed state space perturbations can be expressed. If prior knowledge suggests where decision regions are likely to occur, the centers can be initialized there. Otherwise, we initialize them with $K$-means on the recorded observations so that they begin on the support of the data. We intentionally choose $K$ somewhat large; if some centers are not useful, the sparsity penalty drives their corresponding coefficients in $z_{.,k,.}$ close to zero.

\begin{algorithm}[t]
\caption{Minimum-Perturbation IRL (MP-IRL)}
\label{alg:mpirl}
\begin{algorithmic}[1]
\Require Offline demonstrations $\mathbb{D}=\{(x_{nt},a_{nt},x_{n,t+1})\}_{n=1,t=1}^{N,T_n}$, kernel bandwidth $\sigma$, number of kernels $K$, tradeoff weights $\lambda,\alpha,\beta$, learning rates $\eta_\theta,\eta_\phi,\eta_z,\eta_\mu$
\Require Initial kernel centers $\mu=\{\mu_k\}_{k=1}^K$ \Comment{e.g.\ from prior knowledge or $K$-means on recorded observations}
\Return Base reward parameters $\theta^{(1)}$, final critic parameters $\phi$, kernel centers $\mu$, episode-specific coefficients $\{z_n\}_{n=1}^N$

\State Initialize base reward network $\bar r_\theta$, critic $Q_\phi$, and target critic $Q_{\bar\phi}$
\State Initialize $z_n \gets 0 \in \mathbb{R}^{K\times |\mathcal A|}$ for all $n\in[N]$

\Statex
\State \textbf{Phase 1: Base IRL}
\For{number of phase-1 iterations}
    \State Sample a mini-batch of trajectories from $\mathbb{D}$
    \State Compute $\mathcal L_{\mathrm{base}}(\phi,\theta)$ using Equation~\eqref{eq:mpirl_phase1_compact}
    \State Update $\theta \gets \theta - \eta_\theta \nabla_\theta \mathcal L_{\mathrm{base}}$
    \State Update $\phi \gets \phi - \eta_\phi \nabla_\phi \mathcal L_{\mathrm{base}}$
    \State Update target critic parameters $\bar\phi$
\EndFor
\State Store $(\phi^{(1)},\theta^{(1)}) \gets (\phi,\theta)$

\Statex
\State \textbf{Phase 2: Minimum-Perturbation IRL}
\State Freeze reward parameters at $\theta^{(1)}$ and initialize $\phi \gets \phi^{(1)}$
\For{number of phase-2 iterations}
    \State Sample a mini-batch of trajectories from $\mathbb{D}$
    \State Compute $\mathcal L_{\mathrm{MP}}(\phi,\bm z,\mu)$ using Equation~\eqref{eq:mpirl_phase2_compact}
    \State Update $\phi \gets \phi - \eta_\phi \nabla_\phi \mathcal L_{\mathrm{MP}}$
    \State Update $z_n \gets z_n - \eta_z \nabla_{z_n} \mathcal L_{\mathrm{MP}}$ for trajectories $n$ in the mini-batch
    \If{kernel centers are learnable}
        \State Update $\mu \gets \mu - \eta_\mu \nabla_\mu \mathcal L_{\mathrm{MP}}$
    \EndIf
    \State Update target critic parameters $\bar\phi$
\EndFor

\State \Return $\theta^{(1)}, \phi, \mu, \{z_n\}_{n=1}^N$
\end{algorithmic}
\end{algorithm}

\section{Continuous Navigation Tasks}
\label{app:continuous_navigation}

We use three synthetic continuous navigation tasks. In all three tasks, the recorded observation is the two-dimensional position
\[
x_t = (p_t^x, p_t^y) \in [0,1]^2,
\]
and the action space is
\[
\mathcal{A} = \{\textsc{Right}, \textsc{Left}, \textsc{Up}, \textsc{Down}\}.
\]
Each episode starts from a position sampled uniformly from the bottom-center rectangle
\[
[0.47, 0.53] \times [0.00, 0.05].
\]
We use the term \emph{decision region} for a subset of the observed state space where trajectories that look similar from the recorded observation split because the optimal action depends on an unobserved episode-level context. 

\subsection{Single-Decision Task}
\label{app:single_decision_task}

The single-decision task contains one binary latent variable
\[
z \in \{0,1\},
\]
sampled once at the beginning of the episode and fixed for the whole trajectory. The agent moves in continuous space according to
\[
p_{t+1} = \mathrm{clip}\!\left(p_t + 0.1\, d(a_t) + \varepsilon_t,\ [0,1]^2\right),
\]
where $d(a_t)$ is the unit displacement associated with action $a_t$, $\mathrm{clip}(\cdot,[0,1]^2)$ clips each coordinate to $[0,1]$, and the transition noise is
\[
\varepsilon_t \sim \mathcal{U}([-0.02,0.02]^2).
\]
The maximum episode length is $30$ steps.

Each step incurs a reward of $-0.1$. Let
\[
C = [0.4,0.6] \times [0.4,0.6]
\]
denote the center square. The first time the agent enters $C$, it receives an intermediate reward of $+20$. After this intermediate reward has been collected, the episode terminates with reward $+20$ when the agent reaches the left boundary $p_t^x \leq 0.02$ if $z=0$, or the right boundary $p_t^x \geq 0.98$ if $z=1$.

The unique decision region is the center square $C$. Expert trajectories first move upward toward $C$. Once inside $C$, the hidden context determines whether the agent should move left or right. Therefore, two trajectories can reach nearly the same observed position in the center of the domain but require different optimal actions because they correspond to different values of $z$.

\begin{figure}[t]
    \centering
\includegraphics[width=0.62\linewidth]{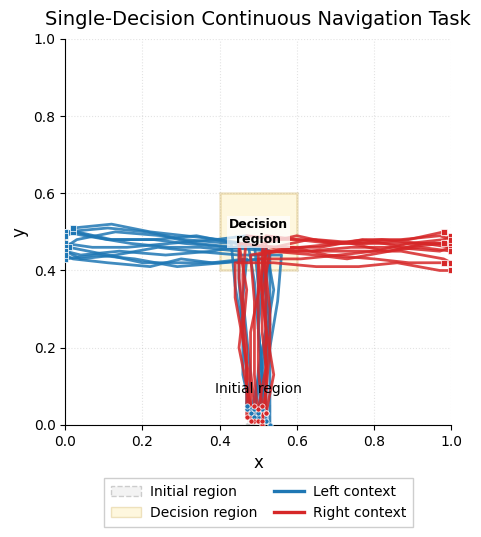}
    \caption{Expert demonstrations for the single-decision task. The center square is the only decision region: trajectories first move upward into this region and then branch left or right depending on the hidden episode-level context.}
    \label{fig:single_decision_demonstrations}
\end{figure}

\subsection{Two-Decision Task (Independent)}
\label{app:two_decision_independent}

The independent two-decision task extends the previous environment by introducing two separate branching points. The hidden context is now
\[
z = (z^{(1)}, z^{(2)}) \in \{0,1\}^2,
\]
where $z^{(1)}$ controls the first decision and $z^{(2)}$ controls the second one. These two latent variables are sampled independently at the start of the episode and remain fixed throughout the trajectory.

The dynamics are
\[
p_{t+1} = \mathrm{clip}\!\left(p_t + 0.1\, d(a_t) + \varepsilon_t,\ [0,1]^2\right),
\]
with smaller transition noise
\[
\varepsilon_t \sim \mathcal{U}([-0.01,0.01]^2),
\]
and the maximum episode length is $50$ steps. Each step incurs reward $-1$.

As before, the first intermediate reward is obtained when the agent first enters the center square
\[
C = [0.4,0.6] \times [0.4,0.6],
\]
which yields reward $+20$. The first left-right decision happens in this region. From there, the agent moves toward one of two side portals:
\[
B_L = [0.0,0.2] \times [0.4,0.6],
\qquad
B_R = [0.8,1.0] \times [0.4,0.6].
\]
If the agent enters $B_L$ and $z^{(1)}=0$, or enters $B_R$ and $z^{(1)}=1$, it receives another reward of $+20$. Upon reaching either side portal, the agent is immediately teleported to the top-center landing region
\[
T = [0.45,0.55] \times [0.90,0.98].
\]

The second decision is made after teleportation. From the landing region $T$, the episode terminates with reward $+100$ at the left boundary if $z^{(2)}=0$, and at the right boundary if $z^{(2)}=1$.

This task therefore has two decision regions: the center square $C$ and the teleport landing region $T$. The key property of this environment is that the two decisions are \emph{independent}. The first latent variable determines whether the agent takes the left or right side portal, while the second latent variable independently determines whether the final destination is the left or right boundary. As a result, all four trajectory types can occur:
\[
(\text{left},\text{left}),\quad
(\text{left},\text{right}),\quad
(\text{right},\text{left}),\quad
(\text{right},\text{right}).
\]

\begin{figure}[t]
    \centering
    \includegraphics[width=0.62\linewidth]{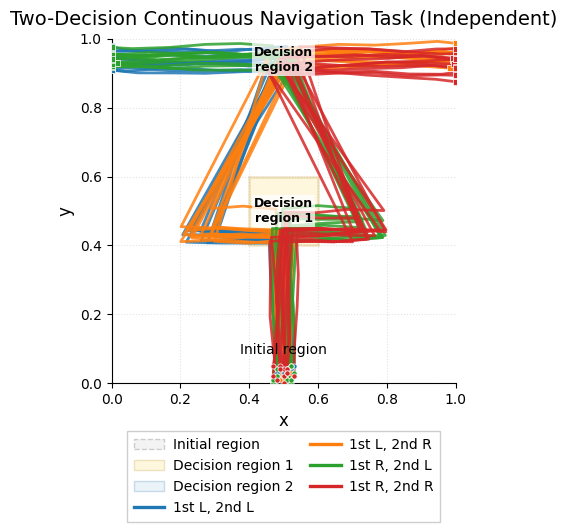}
    \caption{Expert demonstrations for the independent two-decision task. The first decision region is the center square, and the second one is the top-center landing region reached after teleportation. Because the two hidden decisions are independent, all four trajectory families can appear.}
    \label{fig:two_decision_independent_demonstrations}
\end{figure}

\clearpage
\subsubsection{Learned Rewards}
\begin{figure}[h]
    \centering
    \includegraphics[width=.90\textwidth]{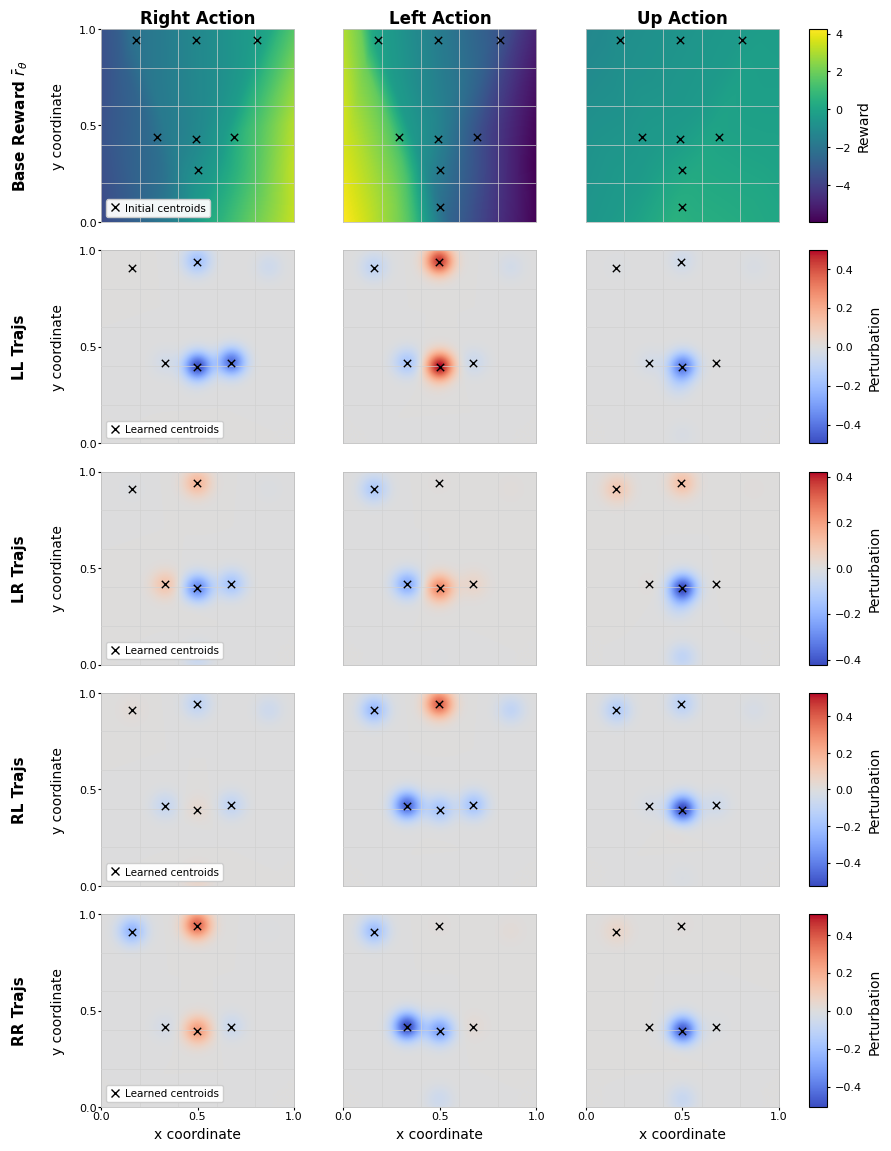}
    \caption{Learned reward structure for the two-decision independent navigation task. The top row shows the base-IRL reward $\bar r_\theta$. The lower rows show the average learned perturbation $z$ for left-left, left-right, right-left, and right-right trajectories across the right, left, and up actions; red indicates positive perturbations and blue negative perturbations. MP-IRL learns distinct perturbation patterns for the four hidden decision contexts.}
    \label{fig:grid_R_ind}
\end{figure}

\clearpage

\subsection{Two-Decision Task (Dependent)}
\label{app:two_decision_dependent}

The dependent two-decision task uses the same geometry, rewards, teleportation mechanism, and transition model as the independent two-decision task. In particular, the agent again moves according to
\[
p_{t+1} = \mathrm{clip}\!\left(p_t + 0.1\, d(a_t) + \varepsilon_t,\ [0,1]^2\right),
\qquad
\varepsilon_t \sim \mathcal{U}([-0.01,0.01]^2),
\]
with horizon $50$, step reward $-1$, center reward $+20$, side-portal reward $+20$, and terminal reward $+100$.

The important difference is that the whole trajectory is now governed by a \emph{single} binary latent variable
\[
z \in \{0,1\},
\]
sampled once per episode. The first decision region is still the center square $C = [0.4,0.6]\times[0.4,0.6]$, and the second decision region is still the teleport landing region $T = [0.45,0.55]\times[0.90,0.98]$. However, the two decisions are no longer independent. Instead, they are tied together by the same hidden context. If $z=0$, the agent should take the left side portal and then finish on the left boundary. If $z=1$, it should take the right side portal and then finish on the right boundary.

Thus, this environment again contains two decision regions, but only two trajectory families are possible:
\[
(\text{left},\text{left})
\qquad \text{and} \qquad
(\text{right},\text{right}).
\]
This makes the dependent task more structured than the independent one: the first decision already predicts the second.

\begin{figure}[t]
    \centering
    \includegraphics[width=0.5\linewidth]{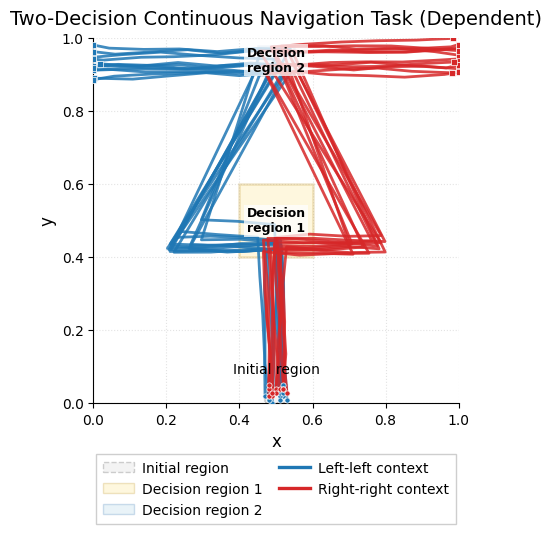}
    \caption{Expert demonstrations for the dependent two-decision task. The two decision regions are the same as in the independent task, but both choices are controlled by the same hidden context, so only the left-left and right-right trajectory families occur.}
    \label{fig:two_decision_dependent_demonstrations}
\end{figure}

\clearpage
\subsubsection{Learnt Rewards}

\begin{figure}[h]
    \centering
    \includegraphics[width=0.90\textwidth]{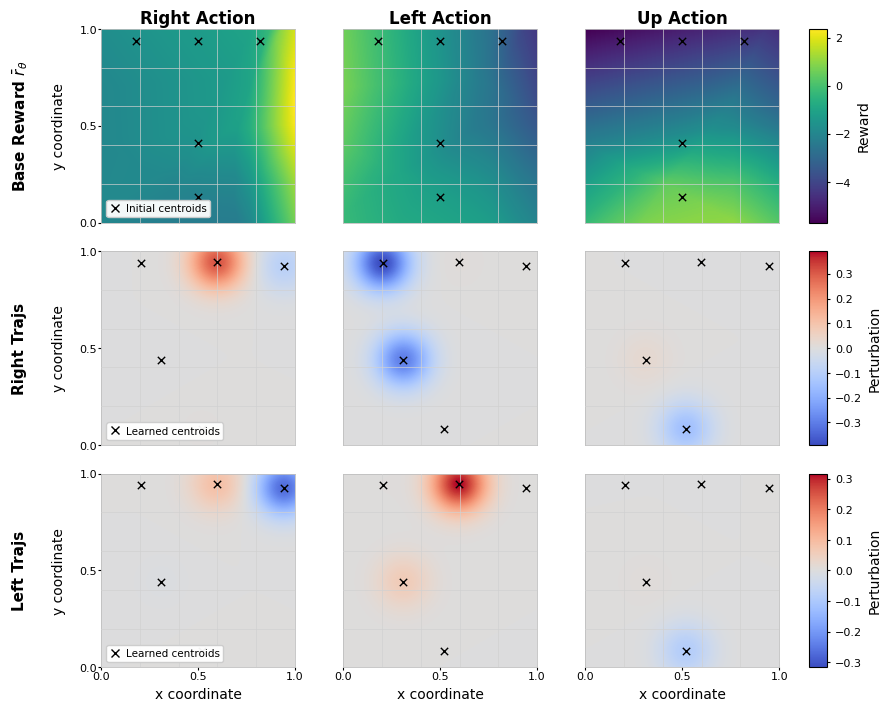}
    \caption{Learned reward structure for the two-decision dependent navigation task. The top row shows the base-IRL reward $\bar r_\theta$. The lower rows show the average learned perturbation $z$ for left-left and right-right trajectories across the right, left, and up actions; red indicates positive perturbations and blue negative perturbations. MP-IRL learns localized perturbations near the two decision regions, with both decisions controlled by the same hidden context.}
    \label{fig:grid_R_dep}
\end{figure}

\section{Cancer Simulator}

\label{app:cancer_simulator}

We also evaluate our method on a cancer treatment simulator based on a low-grade glioma growth model. The environment is a finite-horizon Markov decision process in which each time step corresponds to one month, and the action specifies whether chemotherapy is administered during that month.

The latent physiological state consists of four continuous variables:
\[
s_t = (C_t, P_t, Q_t, Q^p_t),
\]
where $C_t$ is the drug concentration, $P_t$ is the proliferative tissue diameter, $Q_t$ is the quiescent tissue diameter, and $Q^p_t$ is the damaged quiescent tissue diameter. In some experiments, the simulator also appends the current month $t$ to the state, yielding a 5-dimensional state. The action space is binary,
\[
a_t \in \{0,1\},
\]
where $a_t=1$ denotes treatment and $a_t=0$ denotes no treatment. Episodes last for at most $30$ months.

Let
\[
P_t^\star = P_t + Q_t + Q^p_t
\]
denote the total tumor diameter. The simulator evolves according to a discrete-time update of the underlying tumor-growth model. If treatment is given, the drug concentration is first increased by one unit. The state is then updated as
\begin{align}
C_{t+1} &= C_t + a_t - K_{DE} C_t, \\
P_{t+1} &= P_t
+ \lambda_P P_t \left(1 - \frac{P_t^\star}{K}\right)
+ K_{Q_pP} Q^p_t
- K_{PQ} P_t
- \gamma K_{DE} C_{t+1} P_t, \\
Q_{t+1} &= Q_t + K_{PQ} P_t - \gamma K_{DE} C_{t+1} Q_t, \\
Q^p_{t+1} &= Q^p_t + \gamma K_{DE} C_{t+1} Q_t - K_{Q_pP} Q^p_t - \delta_{Q_p} Q^p_t.
\end{align}
Here $K_{DE}$ controls drug decay, $\lambda_P$ is the proliferative growth rate, $K_{PQ}$ is the transition rate from proliferative to quiescent tissue, $K_{Q_pP}$ is the rate at which damaged quiescent tissue returns to the proliferative compartment, $\gamma$ controls treatment efficacy, $\delta_{Q_p}$ is the elimination rate of damaged quiescent tissue, and $K$ is the carrying-capacity parameter. Optionally, multiplicative Gaussian transition noise can be added to the four physiological state variables after the deterministic update.

The reward encourages tumor shrinkage while discouraging excessive treatment. At each month, the reward is
\[
r_t = \left(P_t^\star - P_{t+1}^\star\right) - \eta C_{t+1},
\]
where $\eta$ is a dose-penalty coefficient. Thus, the agent is rewarded for reducing total tumor burden and penalized for large treatment intensity. At the final month, the simulator adds an additional terminal term proportional to the reduction from the initial tumor size:
\[
r_T \leftarrow r_T + \beta \left(P_0^\star - P_T^\star\right),
\]
where $\beta$ is a fixed terminal-reward weight.

Unlike the continuous navigation tasks, the cancer simulator does not contain a localized spatial branching point. A treatment decision is made at every nonterminal step, so every valid state can be viewed as lying in the decision region. In experiments with missing observations, we simulate partial observability by masking selected components of the state before they are given to the learning algorithm, while the simulator itself continues to evolve according to the full latent state.

\begin{figure}[t]
    \centering    \includegraphics[width=\textwidth]{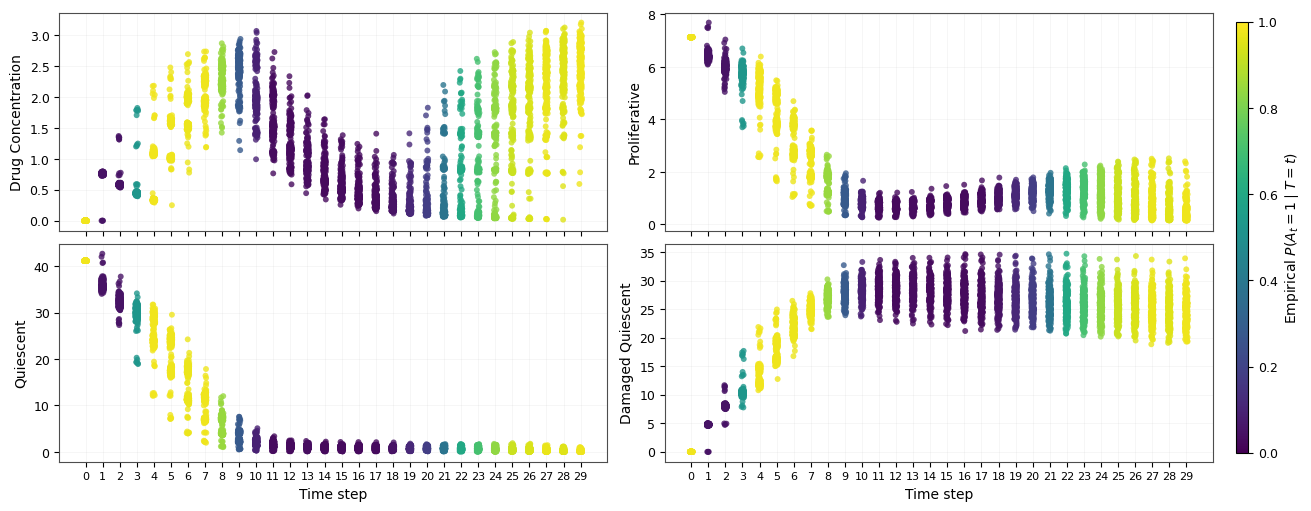}
    \caption{Expert demonstrations in the cancer simulator. Each point is a monthly state from an expert trajectory, colored by the empirical probability of treatment at that time step. The strong time-dependent pattern shows that treatment timing explains much of the behavior, while the physiological variables provide additional context that we selectively mask to create controlled missingness.}
    \label{fig:cancer_demos}
\end{figure}

\section{Real-life ICU Environment}
\label{ICUenvironment}
Following the experiments within a synthetic environments, we now transition to the evaluation of our methodology in a real-world scenario. To this end, we selected the Medical Information Mart for Intensive Care IV (MIMIC-IV) dataset as our experimental field. This dataset offers a rich, diverse, and challenging setting for testing our method, especially given its potential to contribute to advancements in healthcare analytics and patient care strategies.

\subsection{About MIMIC-IV Dataset}

The MIMIC-IV dataset, developed by the MIT Lab for Computational Physiology and publicly available, aggregates a vast range of anonymized health data from critical care units at Beth Israel Deaconess Medical Center in Boston. Covering over a decade's worth of patient admissions, it provides detailed records on demographics, vital signs, lab tests, medications, and more, establishing itself as a critical resource for healthcare model development. Its comprehensive scope spans all patient care aspects, enabling the creation of holistic models for predicting diverse patient outcomes. The dataset's richness lies in its variety, covering over 40,000 patients of different ages, ethnicities, and conditions, and its granularity, offering high-resolution data points and time-stamped records, which are essential for developing precise, dynamic healthcare models. Moreover, MIMIC-IV's public accessibility fosters a global research community's collaboration, enhancing healthcare analytics advancements.

Utilizing the MIMIC-IV dataset, we showcase out the learning applicability of our method in real-world healthcare, to get valuable insights from the data in such a complicated environemnt.

\subsection{Data Preprocessing for Hypotension Analysis}

In our investigation into hypotension within ICU settings, we tailored our preprocessing steps to exclusively include patients affected by this condition. Our methodology commenced with the application of specific filters on the MIMIC-IV dataset to accurately identify the patient cohort of interest. These filters were designed to capture adults aged 18 to 80 years, who had ICU stays of a minimum duration of 24 hours, and exhibited Mean Arterial Pressure (MAP) readings of 65mmHg or below, indicative of acute hypotension.

The analytical framework of this study is structured around a state space defined by a specific set of 15 clinical variables. These variables are categorized into five functional subgroups to facilitate multidimensional analysis:

\begin{table}[h]
\centering
\caption{Categorization of Clinical State Space Variables}
\begin{tabular}{ll}
\toprule
\textbf{Subgroup} & \textbf{Clinical Variables} \\
\midrule
Blood Pressure & Mean, Diastolic, and Systolic Blood Pressure \\
Kidney & Creatinine, Urine Output \\
Vital Signs & Respiratory Rate, Temperature, Heart Rate \\
Oxygen \& Gas & $FiO_2$, $PaO_2$ \\
Perfusion (Other) & Lactate, ALT, AST, GCS \\
Temporal & Time Step \\
\bottomrule
\end{tabular}
\end{table}

Additionally, the variable \texttt{time\_step} is utilized to account for the temporal dimension of the state space.

The action space encompasses two primary treatment modalities: \text{intravenous (IV) fluid bolus therapy} and \text{vasopressor therapy}. This precise filtering approach yielded a dataset comprising 1,684 distinct ICU admissions, from which we derived approximately 100,000 tuples $(\text{state, action, next}\_\text{state}) \in \mathcal{D}$. This dataset serves as the foundation to evaluate our method and the baselines.

\begin{figure}[t]
    \centering    \includegraphics[width=\textwidth]{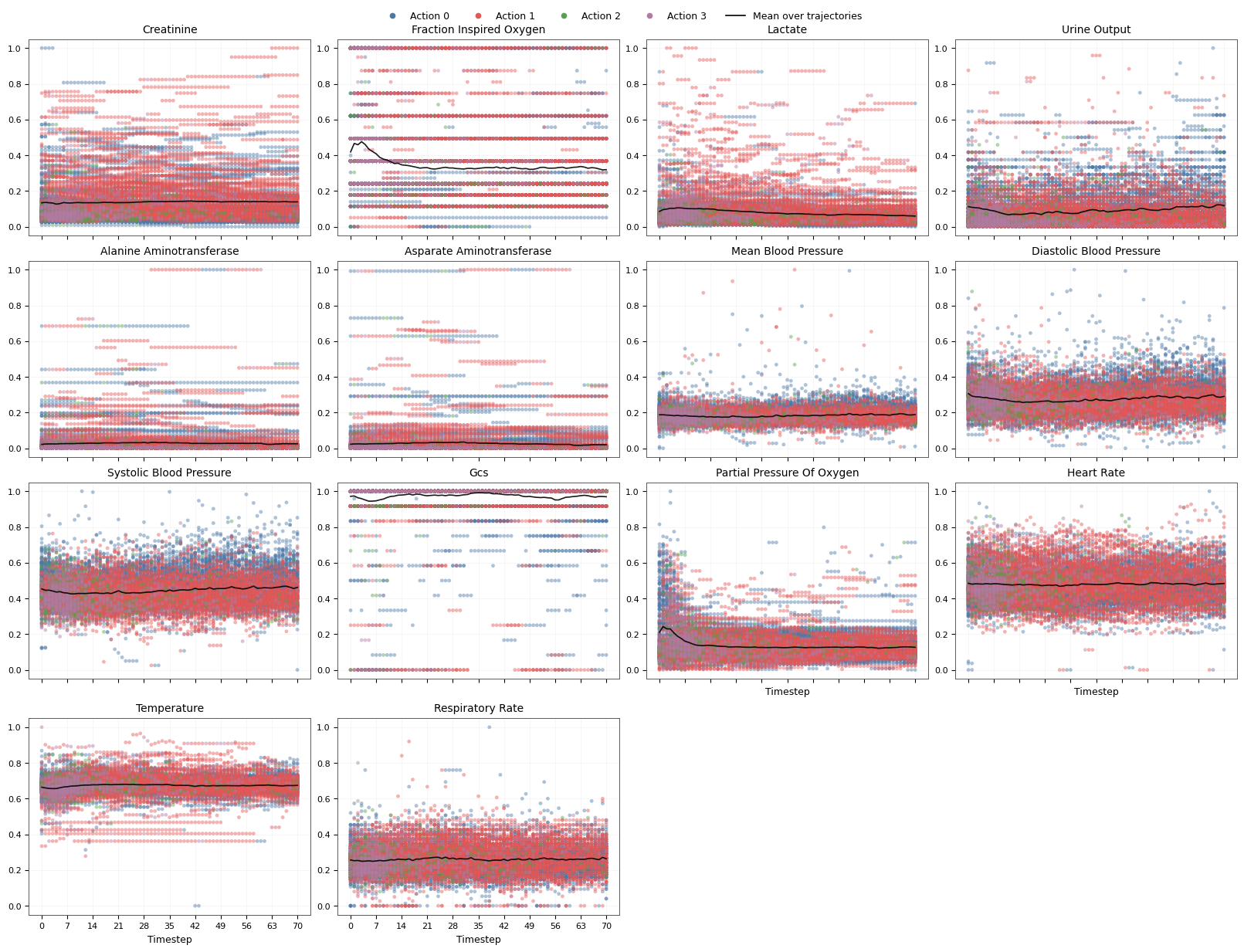}
    \caption{ICU hypotension treatment trajectories from MIMIC-IV. Each trajectory corresponds to one patient stay, with hourly clinical measurements and clinician-prescribed actions. The plots show the temporal evolution of the recorded clinical variables together with the four possible treatment actions: no treatment, vasopressor, IV fluid bolus, or both.}
    \label{fig:mimic_demos}
\end{figure}

\subsection{Action Space Definition}

The action space in our model encapsulates the range of possible treatments administered to patients suffering from hypotension. It consists of four discrete actions, each representing a specific treatment strategy. The actions are enumerated as follows:

\begin{table*}[ht]
\centering
\begin{tabular}{cl}
\toprule
\textbf{Action} & \textbf{Description} \\
\midrule
0 & No treatment administered \\
1 & Vasopressor therapy administered \\
2 & Intravenous (IV) fluid bolus administered \\
3 & Both vasopressor therapy and IV fluid bolus  \\
\bottomrule
\end{tabular}
\caption{Definition of Actions in the Treatment Strategy Space}
\label{table:action_space}
\end{table*}

Each action is designed to reflect the clinical decisions made in the intensive care unit for managing patients' blood pressure levels. Action 0 (no treatment) represents a conservative approach, where no immediate intervention is applied. Action 1 (vasopressor therapy) and Action 2 (IV fluid bolus) correspond to the administration of specific treatments aimed at increasing blood pressure.

\section{Full results}
\label{app:full_results}

\begin{table*}[t]
\vspace{-0.5em}
\centering
\small
\setlength{\tabcolsep}{4.5pt}
\begin{tabular}{llcccc}
\toprule
Setting & Method & Acc. (dec./overall) $\uparrow$ & Acc. (non-dec.) $\uparrow$ & NLL $\downarrow$ & $z$ size $\downarrow$ \\
\midrule
\multicolumn{6}{c}{\textbf{Continuous Navigation}} \\
\midrule
\multirow{2}{*}{Single-decision}
  & IRL    & 67.9  & 99.9 & 183.39 & 0.0 \\
  & MP-IRL & 99.6  & 99.9 & 31.89  & 1.2 \\
\midrule
\multirow{2}{*}{Two-decision dep.}
  & IRL    & 59.6  & 99.8 & 1078.5 & 0.0 \\
  & MP-IRL & 98.4  & 99.9 & 304.9  & 1.1 \\
\midrule
\multirow{2}{*}{Two-decision ind.}
  & IRL    & 59.4  & 99.6 & 1176.7 & 0.0 \\
  & MP-IRL & 97.1  & 99.8 & 269.4  & 2.7 \\
\midrule
\multicolumn{6}{c}{\textbf{Cancer Simulator}} \\
\midrule
\multirow{2}{*}{All features but quiescent}
  & IRL    & 93.73 & --- & 1291.39 & 0.0 \\
  & MP-IRL & 97.68 & --- & 477.43  & 0.54 \\
\midrule
\multirow{2}{*}{Damaged + quiescent + time}
  & IRL    & 91.72 & --- & 1687.03 & 0.0 \\
  & MP-IRL & 97.17 & --- & 834.54  & 0.91 \\
\midrule
\multirow{2}{*}{Time only}
  & IRL    & 87.87 & --- & 1988.29 & 0.0 \\
  & MP-IRL & 96.26 & --- & 1225.75 & 1.65 \\
\midrule
\multicolumn{6}{c}{\textbf{Hypotension Task Management Mimic Dataset}} \\
\midrule
\multirow{2}{*}{All recorded features}
  & IRL    & 63.22 & --- & 12407.45 & 0.0 \\
  & MP-IRL & 89.41 & --- & 5019.83 & 5.63 \\
\midrule
\multirow{2}{*}{Time + low-predictive features}
  & IRL    & 55.24 & --- & 15299.31 & 0.0 \\
  & MP-IRL & 90.66 & --- & 4794.61 & 13.56 \\
\midrule
\multirow{2}{*}{Time only}
  & IRL    & 54.79 & --- & 15372.37 & 0.0 \\
  & MP-IRL & 90.21 & --- & 5032.69 & 13.66 \\
\bottomrule
\end{tabular}
\caption{
For navigation, accuracy is reported inside and outside decision regions; for cancer and ICU, the first accuracy column is overall accuracy. Higher accuracy and lower NLL/$z$ size are better.
}
\label{tab:full_results}
\vspace{1em}
\end{table*}


\newpage
\input{checklist.tex}

\end{document}

%% file: checklist.tex
\section*{NeurIPS Paper Checklist}

\begin{enumerate}

\item {\bf Claims}
    \item[] Question: Do the main claims made in the abstract and introduction accurately reflect the paper's contributions and scope?
    \item[] Answer: \answerYes{} 
    \item[] Justification: The claims are stated in the abstract and introduction and are supported by the experimental results in Section~\ref{sec:experimental_setup} and Section~6.

\item {\bf Limitations}
    \item[] Question: Does the paper discuss the limitations of the work performed by the authors?
    \item[] Answer: \answerYes{} 
    \item[] Justification: Limitations are discussed at the end of Section~7.

\item {\bf Theory assumptions and proofs}
    \item[] Question: For each theoretical result, does the paper provide the full set of assumptions and a complete (and correct) proof?
    \item[] Answer: \answerNA{} 
    \item[] Justification: The paper does not present formal theoretical results or proofs.

    \item {\bf Experimental result reproducibility}
    \item[] Question: Does the paper fully disclose all the information needed to reproduce the main experimental results of the paper to the extent that it affects the main claims and/or conclusions of the paper (regardless of whether the code and data are provided or not)?
    \item[] Answer: \answerYes{} 
    \item[] Justification: We provide algorithmic and experimental details in the main paper and appendix

\item {\bf Open access to data and code}
    \item[] Question: Does the paper provide open access to the data and code, with sufficient instructions to faithfully reproduce the main experimental results, as described in supplemental material?
    \item[] Answer: \answerNo{} 
    \item[] Justification: Code is not released with this submission.

\item {\bf Experimental setting/details}
    \item[] Question: Does the paper specify all the training and test details (e.g., data splits, hyperparameters, how they were chosen, type of optimizer) necessary to understand the results?
    \item[] Answer: \answerYes{} 
    \item[] Justification: Yes, see Appendix

\item {\bf Experiment statistical significance}
    \item[] Question: Does the paper report error bars suitably and correctly defined or other appropriate information about the statistical significance of the experiments?
    \item[] Answer: \answerNA{} 
    \item[] Justification: Our method is more descriptive nature rather than quantitative so we provide results with corresponding visualization on 1 set of demionstrations per experiment

\item {\bf Experiments compute resources}
    \item[] Question: For each experiment, does the paper provide sufficient information on the computer resources (type of compute workers, memory, time of execution) needed to reproduce the experiments?
    \item[] Answer: \answerNA{} 
    \item[] Justification: Only needed my local computer to run experiments
    
\item {\bf Code of ethics}
    \item[] Question: Does the research conducted in the paper conform, in every respect, with the NeurIPS Code of Ethics \url{https://neurips.cc/public/EthicsGuidelines}?
    \item[] Answer: \answerYes{} 
    \item[] Justification: conform with code of ethics

\item {\bf Broader impacts}
    \item[] Question: Does the paper discuss both potential positive societal impacts and negative societal impacts of the work performed?
    \item[] Answer: \answerYes{} 
    \item[] Justification: Yes we talk about how that relate to real clinician settings
    
\item {\bf Safeguards}
    \item[] Question: Does the paper describe safeguards that have been put in place for responsible release of data or models that have a high risk for misuse (e.g., pre-trained language models, image generators, or scraped datasets)?
    \item[] Answer: \answerNA{} 
    \item[] Justification: 

\item {\bf Licenses for existing assets}
    \item[] Question: Are the creators or original owners of assets (e.g., code, data, models), used in the paper, properly credited and are the license and terms of use explicitly mentioned and properly respected?
    \item[] Answer: \answerYes{} 
    \item[] Justification: we credited the dataset used

\item {\bf New assets}
    \item[] Question: Are new assets introduced in the paper well documented and is the documentation provided alongside the assets?
    \item[] Answer: \answerNA{} 
    \item[] Justification: N/A

\item {\bf Crowdsourcing and research with human subjects}
    \item[] Question: For crowdsourcing experiments and research with human subjects, does the paper include the full text of instructions given to participants and screenshots, if applicable, as well as details about compensation (if any)? 
    \item[] Answer: \answerNA{} 
    \item[] Justification: N/A

\item {\bf Institutional review board (IRB) approvals or equivalent for research with human subjects}
    \item[] Question: Does the paper describe potential risks incurred by study participants, whether such risks were disclosed to the subjects, and whether Institutional Review Board (IRB) approvals (or an equivalent approval/review based on the requirements of your country or institution) were obtained?
    \item[] Answer: \answerNA{} 
    \item[] Justification:N/A

\item {\bf Declaration of LLM usage}
    \item[] Question: Does the paper describe the usage of LLMs if it is an important, original, or non-standard component of the core methods in this research? Note that if the LLM is used only for writing, editing, or formatting purposes and does \emph{not} impact the core methodology, scientific rigor, or originality of the research, declaration is not required.
    \item[] Answer: \answerNA{} 
    \item[] Justification: N/A

\end{enumerate}